%% file: main.tex
\documentclass[12pt]{article}

\usepackage[ansinew]{inputenc}
\usepackage[UKenglish]{babel}
\usepackage{authblk}
\usepackage{wrapfig}
\usepackage{textcomp}
\usepackage{mathrsfs}
\usepackage{amsfonts}
\usepackage{mathtools}
\usepackage{amssymb}
\usepackage{amsmath}
\usepackage{amsthm}
\usepackage{amscd}
\usepackage{latexsym}
\usepackage{tensor}
\usepackage{stackengine}
\usepackage{stmaryrd}
\usepackage{rotating}
\usepackage{tikz}
\usetikzlibrary{arrows, calc, decorations.markings, positioning,shapes.geometric}
\usepackage{tikz-cd}
\usepackage{relsize} 
\usepackage[none]{hyphenat}
\usepackage{hyperref}
\usepackage{cleveref}

\usepackage{graphicx}
\graphicspath{ {./Images/} }

\usepackage{hyperref,xcolor}
\usepackage{makeidx}
\definecolor{deepred}{rgb}{0.5,0,0}
\definecolor{deepblue}{rgb}{0,0,0.5}
\definecolor{deepgreen}{rgb}{0,0.5,0}

\hypersetup{
    colorlinks=true,
    linkcolor=deepred,
    filecolor=deepred,      
    urlcolor=deepred,
    citecolor=deepblue,
}

\usepackage{algorithm}
\usepackage{algpseudocode} 

 \usepackage[preprint,nonatbib]{neurips_2026}

\begin{document}

\title{Higher Structures in Deep Learning}
\author{%
	Michael L. Roberts$^{\diamond,}$\thanks{Correspondence to: \texttt{michael@combinatoriallabs.com}}\hspace{3em}
	Carlos Zapata-Carratal\'{a}$^{\S}$\hspace{3em}
	Nicholas J. Cooper$^{\diamond,\dag}$\\
	Lijun Chen$^{\dag}$\hspace{3em}
	Fran\c{c}ois G. Meyer$^{\dag}$\hspace{3em}
	Danna Gurari$^{\dag}$
}
\date{31 May 2026}

\maketitle

\vspace{-0.25em}
\begin{center}
	$^{\diamond}$Combinatorial Labs \hspace{1em} $^{\dag}$University of Colorado Boulder \hspace{1em} $^{\S}$SEMF/Independent Researcher
\end{center}
\vspace{2.25em}

\input{Sections/0-Intro}
\input{Sections/1-RW}
\input{Sections/2-Preliminaries}
\input{Sections/3-Analysis}
\input{Sections/4-Hypernets}
\input{Sections/5-Evolution}
\input{Sections/6-Conclusion}

\bibliographystyle{alphaurl}
\bibliography{references}

\end{document}

%% file: Sections/0-Intro.tex
\begin{abstract}
	We provide an expository introduction on the importance of higher-arity tensor operations to deep learning. Then, we conduct a novel empirical investigation of higher-arity phenomenon in trained neural networks, introduce a hypergraphical generalization of the multilayer perceptron, and explore connections to evolutionary algorithms. We conclude with a discussion of promising directions for future research.
\end{abstract}

\section{Introduction \& Motivation}

Deep learning (DL) has revolutionized many domains in recent years, ranging from language modeling and image generation to reinforcement learning and scientific discovery. Yet, as models become ever more powerful and ubiquitous, their internal mechanics remain opaque. These so-called black box systems offer limited interpretability and minimal insight into the structures that underpin their generalization abilities. Understanding what knowledge models encode, how they encode this knowledge, and how these encodings evolve remain central unresolved challenges.

These challenges arise from a structural problem: deep neural networks are built using ad-hoc combinations of heuristically designed tensor operations, layered onto increasingly large architectures. This approach lacks a principled mathematical foundation. While optimization and empirical performance drive the development of models like transformers, little is known about the underlying formal properties of the tensor interactions that define their function.

This opacity is especially concerning when contrasted with biological systems. In neuroscience, a distinction is made between neural anatomy (the structural layout of connections and regions) and neural activity (the actual responses of neurons over time). Modern DL emphasizes the latter, focusing on the outputs of neurons and layers rather than analyzing the broader compositional structures that dictate model behavior. In doing so, we miss out on opportunities to systematically study how architectural structure itself contributes to learning, reasoning, and generalization.

Tensors, multidimensional generalizations of matrices, lie at the heart of DL. Every model, from convolutional nets to transformers, relies on tensor operations --- e.g., matrix multiplication, convolution, and Hadamard product --- for data representation, manipulation, and learning. Yet, much effort is required to develop a comprehensive mathematical framework for the analysis, generation, and evolution of higher-order tensor interactions in deep neural networks. There is an acute need for the study of these interactions and their corresponding impacts on deep learning.


In this paper, we study tensor interactions in deep neural networks through the lens of higher structures. Here, we use ``higher structures'' to refer to interactions involving $3$ and/or more tensors (higher-arity structure) or interactions involving arrays of dimension $3$ or more (higher-order). The tensor interactions of neural networks are frequently both higher-order and higher-arity, hence our use of the simplifying ``higher structures'' terminology. Leveraging a symbolic and compositional representation of tensor operations, we study the internal representations of deep models, describe a novel generalization of the multilayer perceptron, and discuss connections to evolutionary algorithms. Concretely, we first uncover higher-arity phenomena in logit trajectories in \cref{sec:Analysis}, then develop what we call \textit{neural hypernetworks} in \cref{sec:Generation}, and finally in \cref{sec:Evolution} we discuss how evolutionary algorithms might navigate the combinatorial explosions of higher-order structures.

%% file: Sections/1-RW.tex
\section{Related Work and Contributions}

\subsection{State of the Art of Deep Learning}
Over the last decade, DL has evolved rapidly. Breakthroughs in natural language processing, such as OpenAI's GPT series, Google's PaLM, and Meta's LLaMA models, have demonstrated how impressive performance can be obtained by hyper-scaling models. Similarly, in vision-language models (VLMs) such as CLIP, Flamingo, and Gemini, attention-based architectures have unified disparate modalities under shared representations. These models heavily rely on particular tensor operations, e.g., matrix multiplications in attention layers or tensor convolutions in vision networks.

However, the hyper-scaling status quo shows clear signs of diminishing returns. Improvements in downstream tasks often require exponential increases in parameters, energy, and compute budgets. This highlights an emerging problem: the representational and compositional capacity of existing architectures is nearing a plateau. Deep learning is in need of new theoretical tools to understand these limitations and guide the development of next generation architectures.

The authors' hierarchical framework for neural network architectures \cite{PaperB} provides a foundation for such theoretical tools. Moreover, \cite{PaperB} supports the core hypothesis that higher structures are the key to unlocking powerful new architectures. Specifically, \cite{PaperB} illustrates that the underlying tensor interaction patterns in empirically successful image models do not exploit the full space of compositionality that higher-order tensor structures admit, highlighting the untapped potential of higher-arity tensor operations in deep learning.

\subsection{Theoretical Frameworks for Deep Learning}
Most existing theories, e.g., \cite{GeometricDL, CategoricalDL, NeuroAlgebraicDL, TDLPosition, CopresheafNNs}, are ill-equipped to study the higher structures of deep learning as they restrict their focus to binary tensor operations. Furthermore, the optimization processes used to train deep networks, particularly stochastic gradient descent and its variants, are tightly coupled to the ``binary bias" discussed in \cite{Plexes}. Learning trajectories --- how model parameters evolve in high-dimensional space --- are governed by inner products and matrix operations. These processes, while well-understood in low-arity contexts, become difficult to interpret in higher-arity tensor operation spaces.

This calls for a new paradigm: one that integrates n-ary interactions into the analysis and construction of deep neural networks. Adopting this perspective will not only produce better model architectures (as in \cite{PaperB}), but also grant clearer understanding of training dynamics, representational capacity, and the fundamental principles governing generalization.

\subsection{Contributions}
In this paper, we extend our study of higher-arity tensor operations in deep neural networks by investigating the higher structures hidden in latent representations and introducing the concept of \textit{neural hypernetworks}. We discuss relationships to evolutionary algorithms and provide commentary on promising directions for future research.

%% file: Sections/2-Preliminaries.tex
\section{Fundamentals of Higher-Order Systems}

To understand how ML systems could benefit from higher-order structures, we must look to other domains that already use these concepts. In higher-order systems science, spanning mathematics, physics, and theoretical computer science, researchers have developed formalisms to describe interactions beyond the binary.

One promising concept is the \textbf{hypergraph}, a generalization of graphs where edges can connect more than two nodes. This allows modeling of complex interactions among multiple components simultaneously. Similarly, hypermatrices extend conventional matrices to accommodate more complex indexing and multidimensional relationships.

Current tensor operations in DL are typically reducible to chains of bilinear maps (e.g., matrix-vector products). But many computations in nature are inherently higher-arity. For example, the logical conjunction of multiple facts, or the joint interaction of several features in an image, are not strictly pairwise. Higher-arity algebra, operations involving three or more inputs simultaneously, offers a means to capture these patterns.

We also introduce the concept of the generalized tensor product, extending beyond Kronecker or Hadamard products, to define new forms of interaction between tensors. These generalized products can serve as foundational higher dots, capable of expressing richer computational patterns in model architectures.

Together, these tools form the basis for constructing a new mathematical language of deep learning: one that aligns more closely with the structures we observe in trained models, and potentially in biological cognition.

\subsection{What is Arity?}\label{arity}

The word `arity' \cite{wiki2022arity} is a noun derived from the adjectives `binary', `ternary', `n-ary', etc., typically used to denote the number of elements in a relation or the number of arguments of an operation. Other terms such as `order', `degree', `adicity', `valency', `rank' or `dimension' are sometimes used to refer to similar concepts. We prefer the term `arity' given its emphasis on mathematical compositionality and the fact that it is seldom used outside of abstract algebra and relational structures, in contrast with the aforementioned terms that have well-established uses across mathematical sciences and beyond.

Arity may be generally understood as an elementary quantitative property of the interactions between parts of a system. Let us clarify the notion of arity via an illustrative example. Consider a system of tangled loops of string and the linkedness relations defined on collections of loops by the condition that they stay bound together when each loop is mechanically pulled apart, that is, a physical analogue of mathematical links in knot theory. The smallest linkedness relations are binary since they involve two loops. For instance, the linkedness relation with the simplest topology is known as the Hopf link (Figure \ref{fig:SystemTable}, top). This configuration can be concatenated to form higher-arity relations. Given three loops, we can form the Hopf cycle (Figure \ref{fig:SystemTable}, centre). The condition that three loops form a Hopf cycle is a ternary linkedness relation that is decomposable into pairs of Hopf link relations. We thus say that the Hopf cycle is of arity 3 which is \textbf{reducible} into the lower-arity 2. In contrast, we also find the Borromean link configuration (Figure \ref{fig:SystemTable}, bottom). The condition that three loops form a Borromean link is a ternary linkedness relation that cannot be decomposed into smaller linkedness relations of its constituents. We thus say that the Borromean link is of \textbf{irreducible} arity 3. More generally, loops forming Brunnian links give higher irreducible linkedness relations \cite{baas2014higher}.  The Borromean link is the paradigmatic example of the general phenomenon of \textbf{irreducible arity}: the minimal size of a collection of parts of a system, that determines the nature of their mutual interaction. In the interest of brevity, unless otherwise specified, the term `arity' will mean `irreducible arity'. Thus, we shall refer to the Hopf cycle as a (composite) binary relation and to the Borromean link as a ternary relation.

\subsection{Hypergraphs and Hypermatrices}\label{hyp}

The recent literature on higher-order systems \cite{zhou2006learning,klamt2009hypergraphs,courtney2016generalized,benson2018simplicial,battiston2020networks,battiston2021physics,arsiwalla2021homotopies,arsiwalla2021pregeometric,battiston2022higher} has introduced hypergraphs and simplicial complexes as generalizations of pairwise networks that can account for interactions of arbitrary arity. We shall focus only on \textbf{hypergraphs}, as simplicial complexes are particular cases of hypergraphs endowed with additional structure. Our goal in this section is to demonstrate the adequacy of hypergraphs to model higher-order systems with irreducible interactions and to show that an order-agnostic approach to adjacency properties of hypergraphs leads naturally to higher-arity algebras of hypermatrices.  Consider again the example of a system of tangled loops.

\begin{figure}
	\centering
	\includegraphics[width=0.48\linewidth]{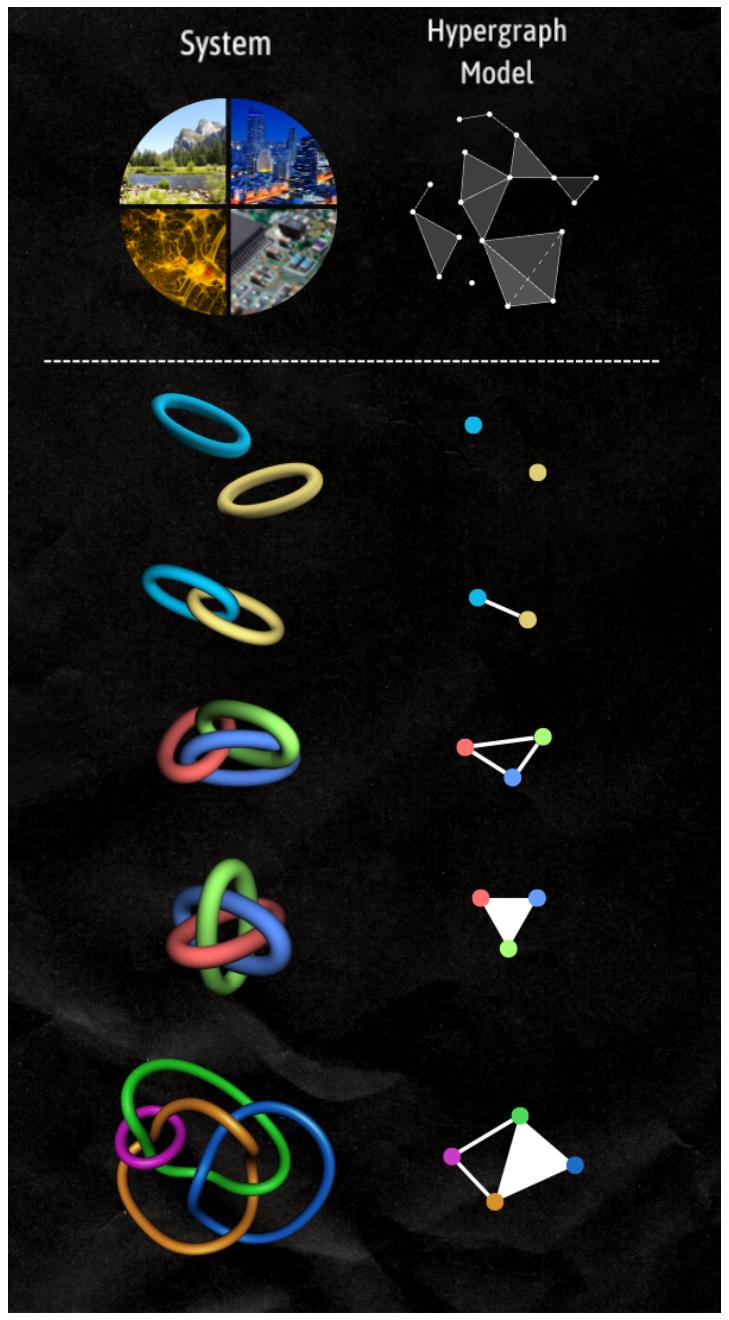} \hspace{1em} \includegraphics[width=0.4825\linewidth]{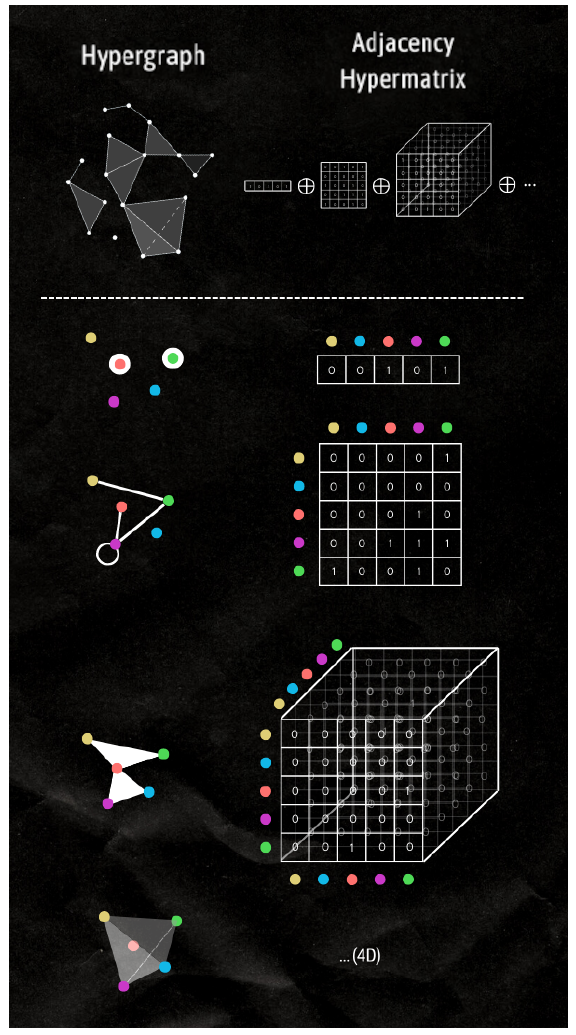}
	\caption{\textbf{Right side} shows tangles of loops, representing general higher-order interactions among parts of a system, are associated with hypergraphs by identifying each loop with a distinct node and each topological link with an (undirected) hyperedge of order equal to the arity of the link. \textbf{Left side} shows a visualization of Boolean-valued adjacency hypermatrices of orders $1$, $2$ and $3$. Higher-order hypermatrices cannot be represented by orthogonal arrays due to the dimensional limitations of 3D space, however they can be regarded as arrays of lower-order arrays.}
	\label{fig:SystemTable}
\end{figure}

Figure \ref{fig:SystemTable} shows how the topology of a tangle can be systematically captured via hypergraph data. Note that the presence of higher-arity links in the tangle demands higher-order hyperedges, a pairwise network model would be insufficient. We can see this explicitly in the different hypergraphs that result from the Hopf cycle and the Borromean link: the former gives a cycle of binary edges while the latter gives a single ternary hyperedge. One of the key advantages of hypergraphs is that they naturally encode pairwise and higher-order data simultaneously, as illustrated by the the bottom tangle in Figure \ref{fig:SystemTable}.

Let us focus on ordinary pairwise networks momentarily. The success of binary graphs as network models can be largely attributed to their capacity to encode \textbf{adjacency} data. At the lowest level, adjacency in binary graphs is simply the condition that nodes share a common edge. More generally, adjacency captures information about the connectivity structure (paths, distances, motifs...) between pairs of nodes. Adjacency data is made operational via the \textbf{adjacency matrix}: by labelling the nodes of a graph with some index variable $i\in [1,N]$ we define the entries of the adjacency matrix $A_{ij}$ according to whether nodes $i$ and $j$ are adjacent. Depending on the kind of graph data one wants to capture, the assignment of adjacency matrix entries can be defined in different ways: the simplest assignment is to take the binary Boolean algebra $(\{0,1\},\vee, \wedge)$ and set $A_{ij}=1$ when there exists at least one edge between nodes $i$ and $j$, and $A_{ij}=0$ otherwise (Figure \ref{fig:SystemTable}, second row); we can also take the natural numbers $(\mathbb{N},+,\cdot)$ and set $A_{ij}$ to be the number of edges that exist between nodes $i$ and $j$; if the graph is weighted over continuous variables, to capture flows or reaction rates for instance, we take the real numbers $(\mathbb{R},+,\cdot)$ and set $A_{ij}$ to be the signed sum of all the weights between nodes $i$ and $j$. In all cases, the entries of the adjacency matrices are elements of a semiring, that is, an abstract set with additive and multiplicative operations $(S,+,\cdot)$ which can be understood as the minimal algebraic setting that enables basic arithmetic.

For general hypergraphs a similar construction is possible if we parse the set of hyperedges by order: for a fixed order $n$, the entry $A^{(n)}_{i_1i_2\cdots i_n}$ is defined according to the existence of a hyperedge between the nodes labelled by $i_1$, $i_2$, $\dots$ $i_n$; this results in a $n$-matrix $A^{(n)}$ that captures the adjacency information at order $n$. The \textbf{adjacency hypermatrix} of a hypergraph is given by the direct sum of all the fixed-order $n$-matrices:
\begin{equation*}
    A:=\bigoplus_{n=1}^\infty A^{(n)}.
\end{equation*}
This construction is illustrated in Figure \ref{fig:SystemTable} for a small set of nodes and Boolean-valued adjacency hypermatrices of order up to $3$.

Although higher-order matrices \cite{cayley1894collected} and relations \cite{peirce1870description,peirce1880algebra} have been known for more than 150 years, non-binary instances of such structures have received very little attention. In fact, higher-arity algebra \cite{dudek2007remarks,azcarraga2010nary,hollings2017wagner,rybolowicz2021topics,zapata2022heaps,rybolowicz2022biunit} and hypermatrix theory \cite{kerner1997cubic,kerner2008ternary,gnang2014combinatorial,gnang2020bhattacharya,gnang2021hypermatrix} are active fields of research that have only started to grow significantly in the last couple of decades, given the state of the art, can only be attained via fundamental mathematical research.

\subsection{Higher-Arity Tensor Operations} \label{higher}

The power of the adjacency matrix -- which is a particular case of adjacency hypermatrix -- resides in its capacity to turn complicated graph topology questions into computationally efficient matrix algebra \cite{estrada2015network}. This is achieved by defining an operation on adjacency matrices from the condition of whether two nodes are connected via some intermediary node. We can generalize this notion to define operations among arbitrary hypermatrices -- commonly referred to as `tensors' in the computer science literature.

The landscape of possible tensor operations is much richer than the classic literature would suggest. Zapata et al. \cite{zapata2024diagrammatic} have given a complete combinatorial description of tensor operations by means of hypergraph diagrams. For the purpose of the present discussion, some basic forms of tensor products will suffice. The \textbf{tensor product} is simply defined by parallel multiplication, i.e. an incidence product with no common indices. For instance, the tensor product of a $1$-array $v$, a $2$-array $m$, and a $3$-array $a$ results in a $6$-array $t$ as follows:
\begin{equation*}
    t(i,j,k,l,n,m):=v(i)\cdot m(j,k)\cdot a(l,n,m).
\end{equation*}

We will also need the notion of a \textbf{generalized tensor product} \cite{GenTenProd}. The setup is the same, namely, each element of the result is determined by `splitting' the indices into the operands. The generalization comes from the realization that one may combine the elements of the operands in different ways. Formally, for a collection of $N$ $V$-valued tensors $a_1, a_2, ..., a_N$ and an $N$-ary operation $\mu: V^N \rightarrow V$, define the \textbf{tensor product over $\mu$}, $\otimes_{\mu}$ as:
\begin{equation*}
    t(i_{1_1}, i_{2_1}, ..., i_{n_1}, ..., i_{n_N}) = \mu(a_1(i_{1_1}, i_{2_1}, ..., i_{n_1}), a_2(i_{1_2}, i_{2_2}, ..., i_{n_2}), ..., a_N(i_{1_N}, i_{2_N}, ..., i_{n_N}))
\end{equation*}
We emphasize that the generalized tensor product used in \cite{GenTenProd} was defined strictly for the binary case. Here, we do not impose such a restriction.

\subsection{Higher Dot Products} \label{dots}

A particularly important class of higher-arity tensor operations are the generalizations of the \textbf{dot product} of component vectors: let two arrays of scalars $\textbf{x} =(x_1,\dots,x_n)$ and $\textbf{y} =(y_1,\dots,y_n)$ then we have the usual definition
\begin{equation*}
    \textbf{x} \cdot \textbf{y} := \sum_{i=1}^n x_i y_i.
\end{equation*}

\begin{wrapfigure}[29]{r}{0.475\textwidth}
	\centering
	\raisebox{0pt}[\dimexpr\height-0.0\baselineskip\relax]{\includegraphics[width=0.95\linewidth]{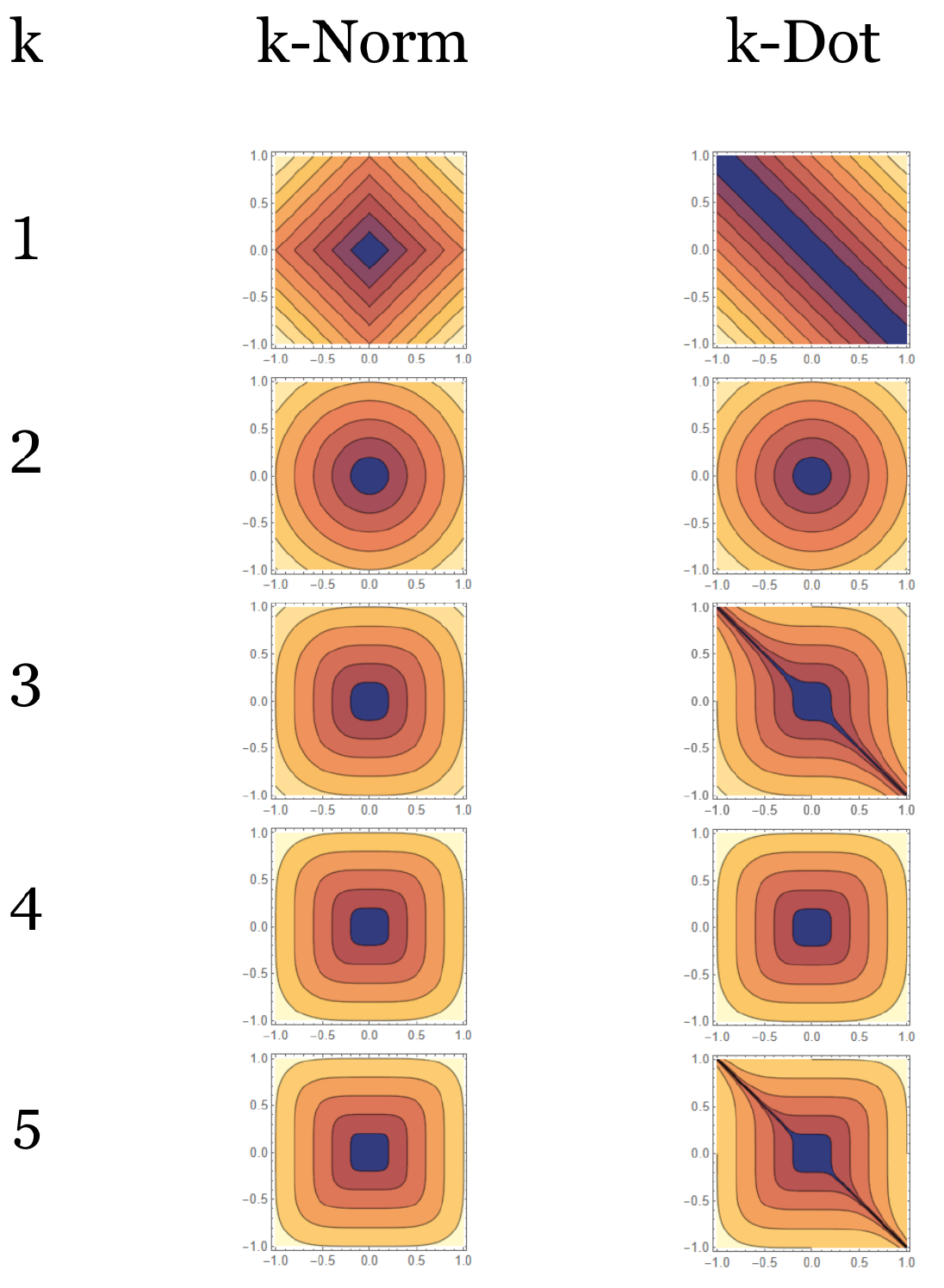}}
	\caption{A comparison of ordinary norms and higher-arity dots. We can observe the discrepancy of odd arity norms and dots. This may result in better analytical tools for classification tasks that mix distance and dichotomy.}
	\label{fig:k-Dots}
\end{wrapfigure}
From this definition, we derive important notions such as the \textbf{Euclidean distance} or the \textbf{cosine similarity}, which are widely used in data analysis. One can easily form a ternary analogue of the dot product wherein three component vectors are operated to produce a single scalar:
\begin{equation*}
    \cdot (\textbf{x}, \textbf{y}, \textbf{z}) := \sum_{i=1}^n x_i y_iz_i.
\end{equation*}
In general, we define the \textbf{$k$-ary dot product} of vectors of $n$ scalar components as the quantity:
\begin{equation*}
    \cdot (\textbf{x}^{(1)}, \textbf{x}^{(2)}, \dots, \textbf{x}^{(k)}) := \sum_{i=1}^n \prod_{j=1}^k x_i^{(j)}.
\end{equation*}
The seeming simplicity of this generalization hides the fact that very little mathematical theory is known about dot products for $k>2$. They offer powerful computational devices but are hard to interpret geometrically, unlike the case of $k=2$ where the geometric interpretation is clear -- indeed, that case recovers Euclidean geometry.

To illustrate how higher dot products extend ordinary geometry and potentially provide a novel way to define distance-like metrics we define the \textbf{$k$-dot} of a component vector $\textbf{x} =(x_1,\dots,x_n)$ simply as:
\begin{equation*}
    D_k(\textbf{x}):=\sqrt[k]{|\cdot (\textbf{x}, \textbf{x}, \dots, \textbf{x})|}.
\end{equation*}
This contrast with the ordinary $L_k$ norms that are defined from the absolute value (the natural norm on the real numbers) as:
\begin{equation*}
    \|\textbf{x}\|_{k}:=\sqrt[k]{|x_1|^k + |x_2|^k + \dots + |x_n|^k}.
\end{equation*}
Note that both notions coincide for even arity, as it is indeed easy to verify:
\begin{equation*}
    D_k(\textbf{x})=\|\textbf{x}\|_{k} \qquad \text{for } k=2l, l\in \mathbb{N}.
\end{equation*}
The behavior of the higher dots in odd arities can be seen in Figure \ref{fig:k-Dots}.

%% file: Sections/3-Analysis.tex
\section{Higher-Order Analyses of Deep Models\label{sec:Analysis}}

We now employ higher dot-products to gain insights into deep neural network \textit{representations} over \textit{training trajectories}. To accomplish this, we develop an \textbf{arity filtration} algorithm, then use it to study the higher order structures hidden in DNNs. We will see that higher order similarity measures on logit spaces exhibit surprising `spiking' behavior which corresponds to the onset of overfitting in the training process, a trend which generalizes across datasets and model architectures. Furthermore, this phenomenon is \textit{not} present in binary similarity measures, suggesting that unexplored `emergent behavior' lies hidden in the higher order structures of neural representations.

\subsection{Motivation}
Binary similarity (dot product) has proved to be a valuable tool for understanding how NNs process data \cite{RegularizationAndGenVsID, SemNC, IntermediateNeuralCollapse, SlicedMI, PerLayerCompDisc}, which begs the question: \textit{what higher order structure lies hidden in NN representations?} Towards answering this question, we systematically analyze the higher order similarities of neural representations. However, naively computing higher-arity similarity tensors proves to be intractable for real world data due to exponential memory cost. This necessitates the creation of an efficient filtration algorithm. 

\subsection{Arity Filtrations}
Consider a set of $N$ $D$-dimensional points $\mathbf{X} \in \mathbb{R}^{N \times D}$. We define a \textit{cell} to be subset of data points. We call the cardinality of a cell its arity. Let $C_{\alpha}$ denote the set of all $\alpha$-ary cells. Define a \textit{cell multiplication function} $\mu_{cell}$ by the following rule:
\begin{equation*}
    \mu_{cell}(c_1, c_2, ..., c_N) = \sum_{i = 1}^{D} \prod_{j = 1}^{N} c_1^1[i] \cdot c_1^2[i] \cdot ... \cdot c_1^{|c_1|}[i] ... c_N^1[i] \cdot ... \cdot c_N^{|c_N|}[i]
\end{equation*}
In other words, $\mu_{cell}$ computes the arbitrarily high-arity dot product between all elements of all cells.

The filtration algorithm begins by computing the standard dot product similarity matrix, that is, an order 2 tensor containing all pair-wise similarities. Said another way, we begin with the tensor product over $\mu_{cell}$ of the set of all unary cells ($C_1$) with itself. This naturally provides similarity scores for all $\binom{N}{2}$ binary cells. To inductively ascend the arity ladder, we select the top-$N$ binary cells ($C_2^N \subsetneq C_2$) with the highest similarity scores. This gives another set of $N$ cells with which we can compute a tensor product over $\mu_{cell}$, yielding similarity scores for a subset of $C_3 \cup C_4$, the top-$N$ of which are selected, and so on. This process continues until there are no cell similarity scores above a user defined threshold. The algorithm is described in detail in \textbf{Algorithm \ref{alg:arityfilt}}.

\begin{algorithm}
\caption{Arity Filtration}\label{alg:arityfilt}
\begin{algorithmic}
\Require $\mathbf{X} \in \mathbb{R}^{N \times D}$ \Comment{Dataset}
\Require $\tau > 0$ \Comment{Threshold}
\State $C \gets \{\{\mathbf{X}[i,:]\} \text{ } : \text{ } 1 \leq i \leq N\}$ \Comment{Form the set of all unary cells}
\State $R \gets list[\emptyset, \emptyset, ..., \emptyset]$ \Comment{Storage for results}
\While{$|C| > 1$}
\State $S \gets C \otimes_{\mu_{cell}} C$ \Comment{Compute the cell tensor product}
\State $S \gets \{S[i,j] \text{ } : \text{ } (i,j) \in N\times N, S[i,j] > \tau\}$ \Comment{Discard similarities below the threshold}
\For{$(i > j) \in N\times N$}
\State $\alpha \gets |C_i \cup C_j|$
\State $R[\alpha] \gets R[\alpha] \cup S[i,j]$ \Comment{Keep track of the similarity scores per arity}
\EndFor
\State $I \gets \{argsort(S)[i] \text{ } : \text{ } 1 \leq i \leq N\}$ \Comment{Get the indices of the top-$N$ similarity scores}
\State $C \gets \{\{C_i \cup C_j\} \text{ } : \text{ } (i,j) \in I\}$ \Comment{Merge the cells with the highest similarities}
\EndWhile
\State Return $R$
\end{algorithmic}
\end{algorithm}

\paragraph{Computational Cost} These steps of interleaved `matrix multiplication' and top-$N$ cell selection allow increasingly high-arity similarities to be computed while avoiding the prohibitive $N^{\alpha}$ memory costs when $\alpha > 2$. In fact, the computational cost for each `level' of the filtration is approximately the same as the initial binary similarity matrix. Moreover, the memory cost to store the results of each `level' of the filtration is just a constant $N^2$, which remains tractable even for reasonably large $N$. We were able to process data up to $N\approx1k-10k$ on a dual Nvidia RTX 3090 equipped workstation with 128GB of memory.

\paragraph{Limitations} Of course, the computational benefits of this filtration algorithm come at a price: not all possible similarity scores are computed for $\alpha > 2$. However, we will demonstrate that when the data are normalized appropriately, the top-$N$ sampling scheme is provably optimal in the sense that only `insignificant' similarities are discarded.

\subsection{The Question of Normalization}
The top-$N$ cell sampling makes a strong implicit assumption: low similarity score at low-arity implies low similarity score at higher-arity. In other words, if $\alpha$ vectors have a low $\alpha$-dot similarity, then adding another vector can't suddenly result in a high $\alpha + 1$-dot similarity. Yet, this intuition is, in general, completely false. Consider the normal binary dot-product between the vectors $\mathbf{x} = (1, 0.1, 0.1)$ and $\mathbf{y} = (0.1, 1, -1)$. Of course, $\mathbf{x} \cdot \mathbf{y} = 0.1$. Next, consider the ternary dot-product of $\mathbf{x}, \mathbf{y}$, and $\mathbf{z} = (1, 1, -1)$. We notice that $\cdot(\mathbf{x}, \mathbf{y}, \mathbf{z}) = 0.3 = 3(\mathbf{x} \cdot \mathbf{y})$. This `emergent' property has a nice geometric interpretation: given some collection of vectors, one can usually find another vector which increases the higher-arity similarity by considering the \textit{orthant signature}. That is to say, higher-arity similarities capture some notion of `orthant alignment' between the vectors.

This phenomenon is bad news for our arity filtration algorithm. How can we conduct analysis on the higher-arity similarity distributions when the top-$N$ sampling might discard `emergent' structures? The answer lies in discretization. Specifically, we consider a binary $0/1$ discretization, that is, negative values are sent to zero, and positive values are sent to one. In this discrete setting, we have the following:
\begin{equation*}
    \forall \alpha \geq 2,
    \cdot(\mathbf{x}_1, \mathbf{x}_2, ..., \mathbf{x}_{\alpha}) = 0 \implies \cdot(\mathbf{x}_1, \mathbf{x}_2, ..., \mathbf{x}_{\alpha}, \mathbf{x}_{\alpha + 1}) = 0, \forall \mathbf{x}_{\alpha + 1}
\end{equation*}
Moreover, a stronger statement is true:
\begin{equation*}
    \forall \alpha \geq 2,
    \cdot(\mathbf{x}_1, \mathbf{x}_2, ..., \mathbf{x}_{\alpha}) \geq \cdot(\mathbf{x}_1, \mathbf{x}_2, ..., \mathbf{x}_{\alpha}, \mathbf{x}_{\alpha + 1}), \forall \mathbf{x}_{\alpha + 1}
\end{equation*}
This can be easily seen by taking $\mathbf{x}_{\alpha + 1}$ as the vector of all ones. The above inequality guarantees that the top-$N$ sampling employed in the arity filtration algorithm selects only the cells which are `most likely' to produce meaningful higher-arity similarity scores.

While this discretization may appear to be a brutish hack, it has a natural and meaningful interpretation in the context of neural representations. Most traditional neural networks process data by sequentially projecting the points onto sets of hyperplanes, and as such, the binary discretization can be thought of as a summary of which hyperplanes the data `agrees with'. Alternatively, one can think of each string of $0$s and $1$s as an orthant signature, describing which region of high dimensional space the data reside in. For this reason, we will sometimes refer to this binary discretization as \textit{orthant normalization}; each data point is replaced with a `canonical' representative of its corresponding orthant. It is worth mentioning that the popular and effective ReLU family of activation functions \footnote{Such as Leaky ReLU, GELU, etc...} implicitly enforce a milder version of orthant normalization, further justifying our use of it for the analysis of neural representations. 

It is easiest to reason about this orthant normalization in context of \textit{logits}. In classification models, logits are the pre-activation values at the final output layer. Typically, they are converted into class scores via the softmax function to produce a probability distribution over the number of classes defined by the classification task at hand. It is well understood that logits contain a wealth of information about the `dark knowledge' encoded by deep neural networks \cite{SKD, LogitStandardizationKD}, making them a natural candidate for analysis via our arity filtration algorithm. In the context of logits, orthant normalization can be interpreted as a summary of the plausible class predictions for a data point, with $1$ indicating the point `looked like' the class corresponding to one of the hyperplanes in the final linear layer. For example, the string $(1, 1, 0, 0, 1, 0)$ could encode the statement `it looks like a cat, but also like a dog and a cow, yet it definitely doesn't look like a giraffe, car, or bus'.


\subsection{Analysis of Training Trajectories}
At last, we arrive at the point: using orthant normalized arity filtrations to study the training process of DNNs. We consider a simple image classification setting with both CNNs and Vision Transformers on standard `toy' datasets. We compute the logit scores for a (class balanced) set of $5000$ points, perform orthant normalization, then run \textbf{Algorithm \ref{alg:arityfilt}} with $\tau = 3$. This enables the computation of statistics of the similarity scores at various arities, specifically, we consider $\alpha \in \{2, 4, 6, 8, 10, 12\}$. Intuitively, this process captures the `depth' to which different points `look the same' in terms of their plausible classes.

\begin{figure}
    \centering
    \includegraphics[width=\linewidth]{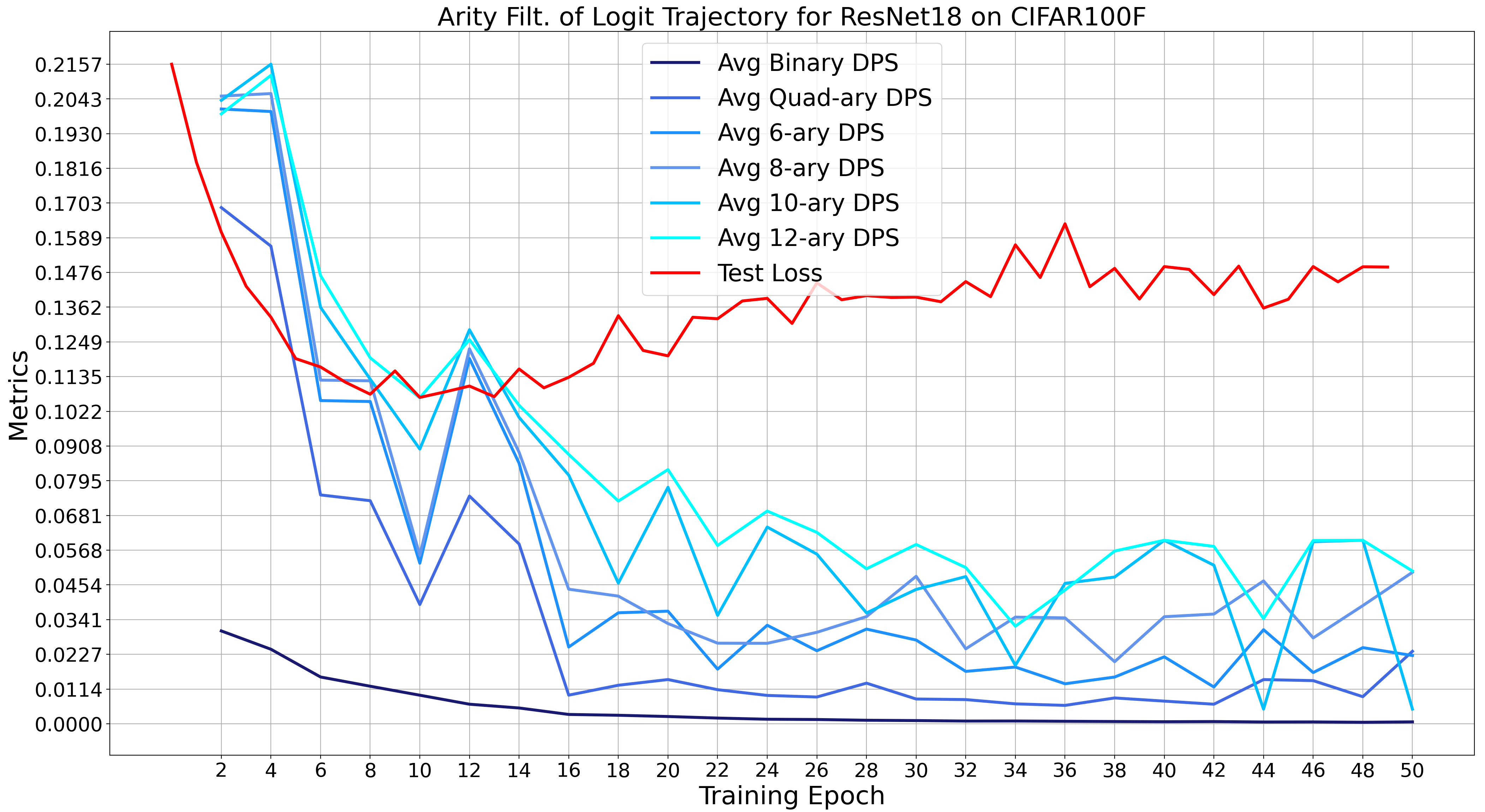}
    \includegraphics[width=\linewidth]{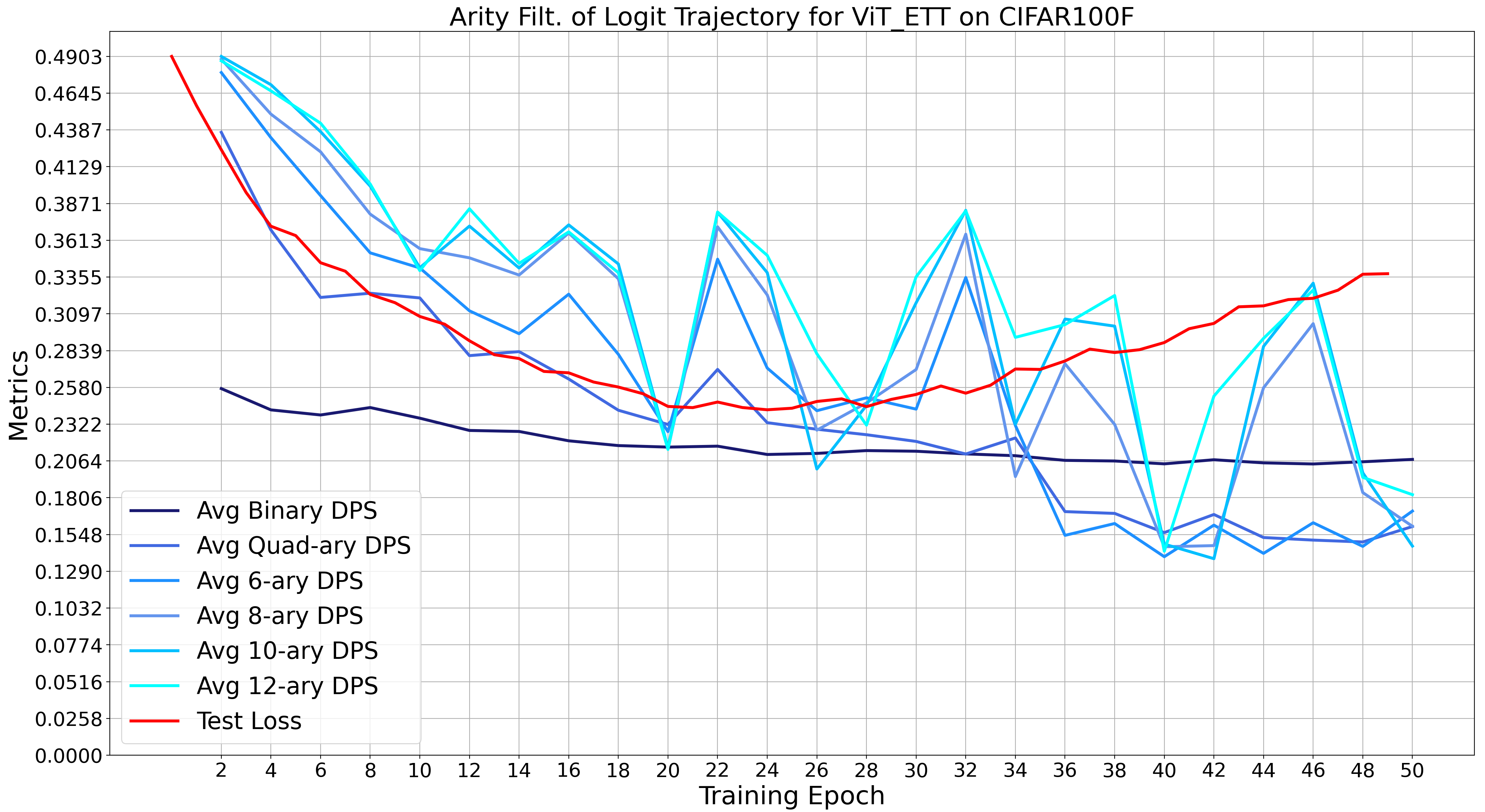}
    \caption{Arity Filtration results on CIFAR100, Top: ResNet18, Bottom: ViT\_ETT. Average $\alpha$-dot similarity scores shown in shades of blue, test loss shown in red and rescaled for legibility. Patterns emerge in similarity scores with $\alpha > 2$ which correspond to model overfitting.}
    \label{fig:ArityFilt_C100}
\end{figure}

\begin{figure}
    \centering
    \includegraphics[width=\linewidth]{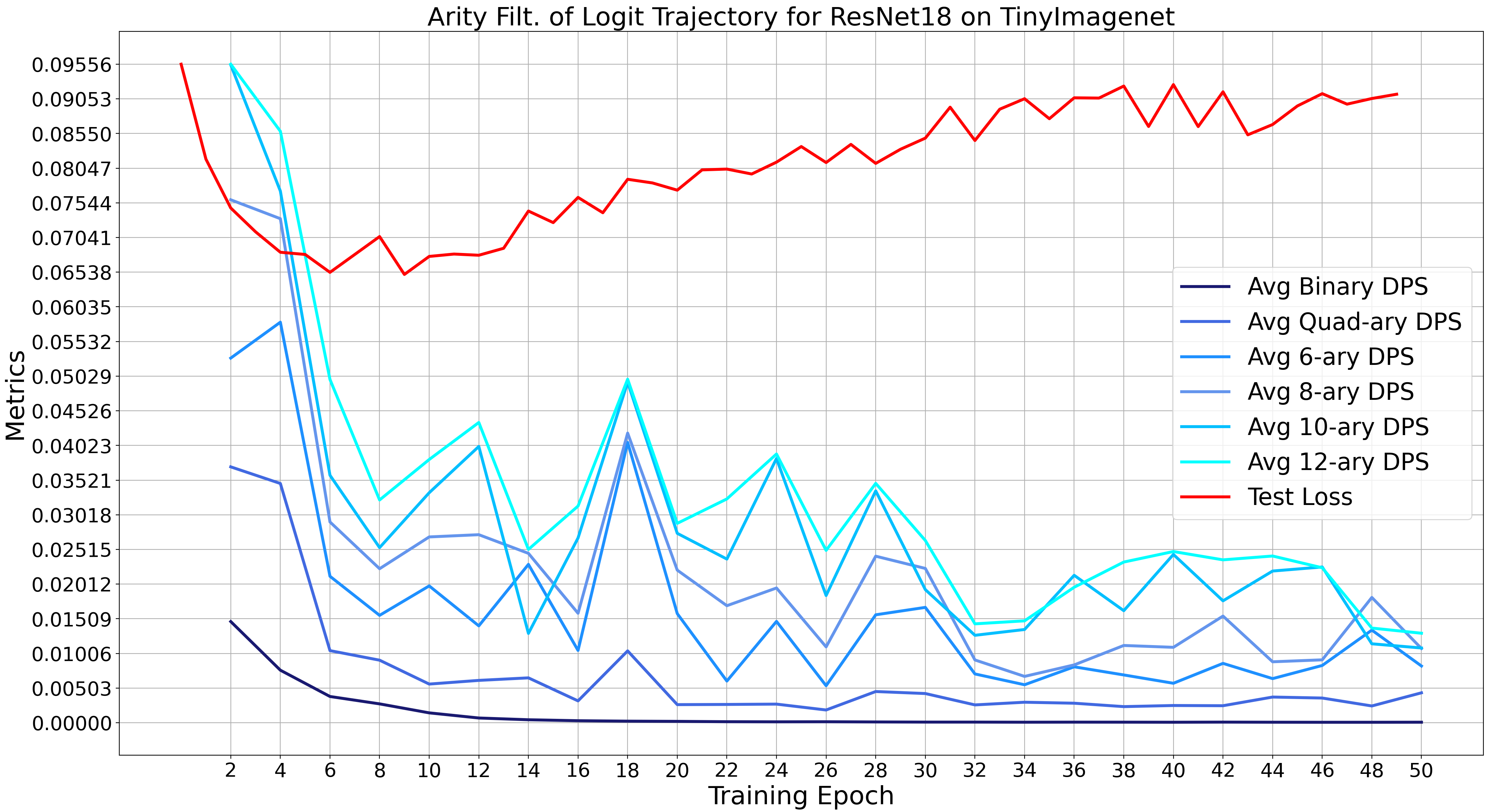}
    \includegraphics[width=\linewidth]{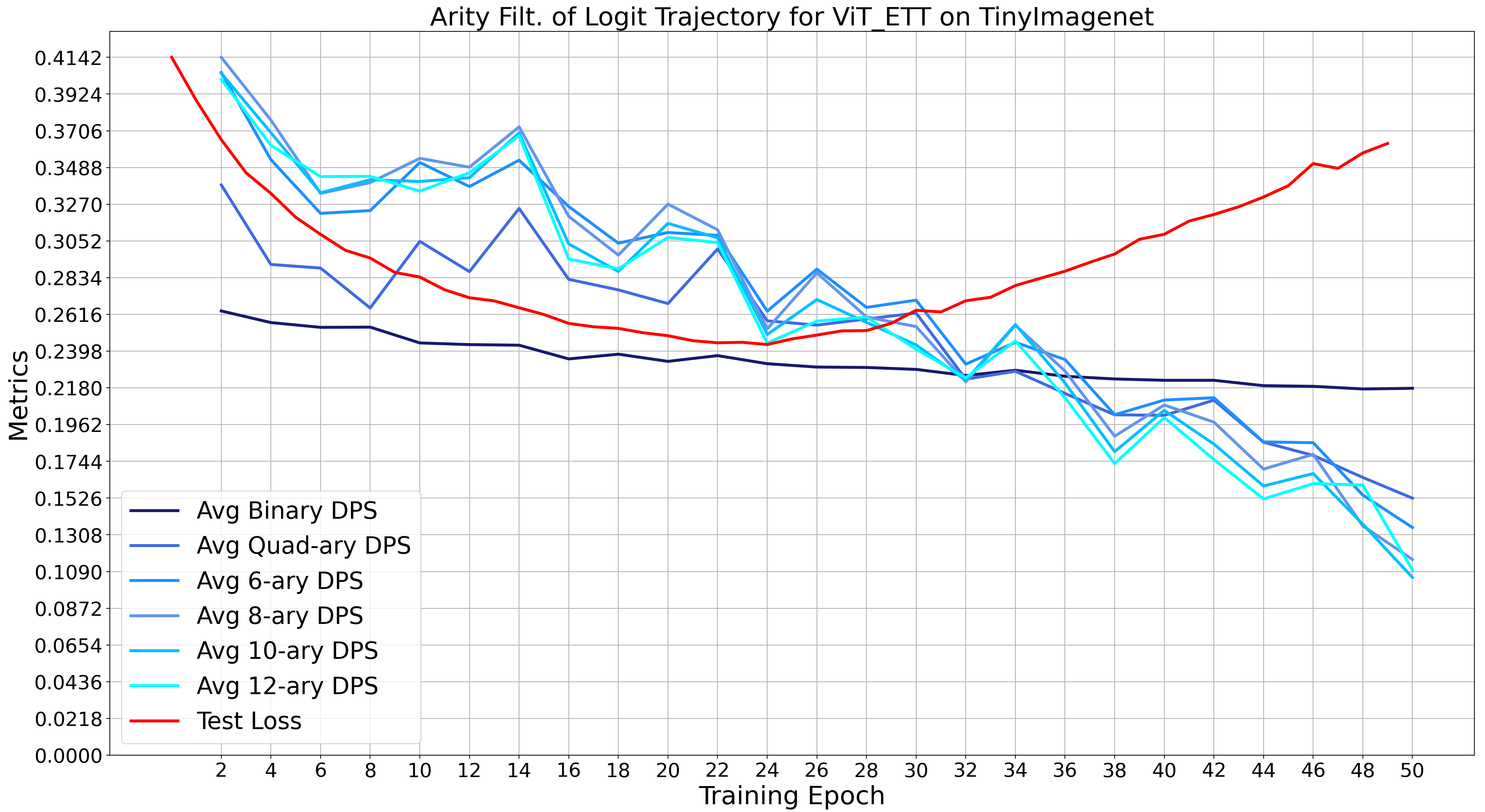}
    \caption{Arity Filtration results on TinyImageNet, Top: ResNet18, Bottom: ViT\_ETT. Average $\alpha$-dot similarity scores shown in shades of blue, test loss shown in red and rescaled for legibility. Patterns emerge in similarity scores with $\alpha > 2$ which correspond to model overfitting.}
    \label{fig:ArityFilt_TIM}
\end{figure}

There is growing interest in understanding the behavior of DNNs across the \textit{training trajectory} \cite{TDATraj}, meaning the evolution of structural attributes as the training process progresses. Accordingly, we conduct the arity filtration analysis across the training process. Results are shown in \textbf{Figures \ref{fig:ArityFilt_C100} and \ref{fig:ArityFilt_TIM}}.

\paragraph{Datasets} We experiment with the CIFAR100 \cite{CIFAR} and TinyImageNet \cite{TIM} datasets. CIFAR100 consists of $60,000$ $32\times32$ color images distributed across $100$ classes, while TinyImageNet consists of $110,000$ $64\times64$ color images distributed across $200$ classes. These datasets provide two distinct learning problem scales, while remaining small enough to permit computationally intensive analysis.

\paragraph{Models} We use ResNet18 \cite{ResNet} and ViT\_ETT (extra tiny) \cite{ViT} for the arity filtration experiments to demonstrate that our findings apply to both CNNs and Vision Transformers. ResNet18 has $\approx 11M$ parameters, while ViT\_ETT has $\approx 3M$. These models were chosen as minimal size representative examples of popular classes of architecture.

\subsection{Results}
A surprising trend emerges from the arity filtration results: model overfitting is detectable by examining $(\alpha > 2)$-dot similarity scores! Specifically, when the test loss begins to increase we observe dramatic spikes in the higher-arity similarity scores which are not present in the binary case. For example, consider ResNet18 on CIFAR100 (top of figure \ref{fig:ArityFilt_C100}). Model overfitting begins in earnest around epoch 15, immediately before which all $(\alpha > 2)$-dot scores increase dramatically for a few epochs. Similar behavior is observed with the vision transformer on CIFAR100, however, several spikes are present instead of a single spike. Continuing this pattern, ResNet18 on TinyImageNet produces a `weaker' spike at the beginning of the overfitting phase followed by several `stronger' spikes. Here, `weaker' means only very high ($\geq 10$) arity similarity scores exhibit the spiking behavior.

These results suggest that as models begin to overfit, they dramatically reorganize the higher order similarity structure of their logit scores. In other words, the hyperplanes learned by the final classification layer begin to move chaotically with respect to the extracted representations around the same time as overfitting starts. This highlights the importance of considering higher order similarity measures in deep neural networks, as this discovery could not have been made by considering only the binary dot-product similarity measure. In turn, these results lead us to an intriguing question: are modern DNNs limited by the restriction to sequential binary similarity and non-linear activation?

\subsection{Summary}
In this section, we introduced an arity filtration algorithm which enables the tractable computation of arbitrarily high-arity similarity measures on point cloud data. We demonstrated that `spiking' behavior emerges in the higher order similarities of DNN logits in close correspondence with the start of model overfitting. While these preliminary results provide evidence of the promise of incorporating higher order mathematics into the analysis and design of DNNs, many fascinating research questions remain open. We highlight some possible directions here. (1) Can a sampling scheme be developed such as to enable the analysis of higher similarities in the absence of discretization? (2) Can arity filtration analysis be extended to latent representations? If so, what else lies hidden behind the $\alpha > 2$ curtain? (3) Do these results extend to larger scale models and datasets? What about other tasks such as image segmentation, language modeling, etc? (4) How might higher similarities be incorporated into the design of next generation DNNs? In the next section, we provide a constructive conjecture on how one might approach this final question. 

Beyond the technical details, the results shown in this section firmly establishes the central thesis of this paper: ordinary DNNs display higher-order phenomena -- and there is no reason to believe this is architecture-specific. Therefore, we have laid out the starting point of a line of research that demands mathematical tools that can account for the higher-order interactions to better model multiple aspects of ML systems.

%% file: Sections/4-Hypernets.tex
\section{Constructing New Deep Architectures\label{sec:Generation}}

Building on our formalism, we now turn to the design of new DL architectures. One of the most compelling applications of hypergraphs is as a blueprint for artificial neural anatomy. Just as the brain features interconnected regions supporting multi-way communication, hypergraph-based models can instantiate higher-arity pathways between neurons or modules.

We propose a procedure for generating feedforward networks from hypergraphs, where each hyperedge defines a composite operation involving multiple inputs. These can be mapped onto computational units that perform generalized tensor contractions, resulting in architectures that support inherently multi-way processing.

Such models are not simply deeper or wider, but structurally richer. We hypothesize that by incorporating higher-arity tensor operations directly into the architecture, we might create models that learn more efficiently, generalize better, and encode knowledge in more semantically coherent ways.

Early implementations of these ideas can leverage existing DL frameworks with minimal adaptation, by defining custom modules or layers that perform the required higher-order operations. In the long term, however, we anticipate the development of new compiler and hardware tools to natively support these operations.

This approach shifts the focus of model design from heuristics and trial-and-error to principled mathematical construction, potentially leading to a new generation of models with capabilities shaped by their intrinsic structural properties.

\subsection{Higher-Arity Feedforward Networks} \label{haffn}

The higher tensor operations introduced in Section \ref{higher} inspire a direct generalization of the usual feedforward network architecture commonly found in many elementary components of DL models. In this section we introduce the novel notion of \textbf{generalized feedforward neural hypernetwork} (NHN) as a promising candidate of a class of new model architectures that fundamentally integrate higher-order structures.

The basic layout is similar to conventional feedforward networks: a series of nodes are distributed in a discrete number of layers, where we think of the leftmost nodes as the input layer and the rightmost ones as the output layer. For instance, here would be a small example:
\begin{center}
    \includegraphics[scale=0.5]{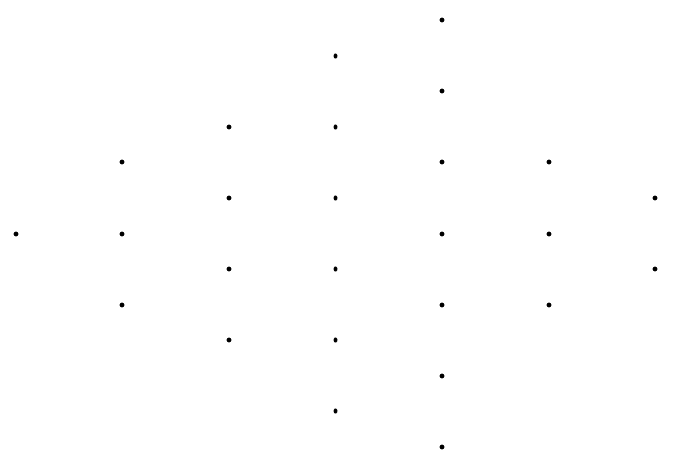}
\end{center}
In an ordinary feedforward network, a series of pair-wise edges (with accompanying weights) will be distributed between adjacent layers. Here we generalize this notion by allowing hyperedges more generally, that take several node values at one given layer and operate them into a single node of the subsequent layer. In our small example, this would be a typical sparse distribution of hyperedges of a generalized feedforward neural hypernetwork:
\begin{center}
    \includegraphics[scale=0.5]{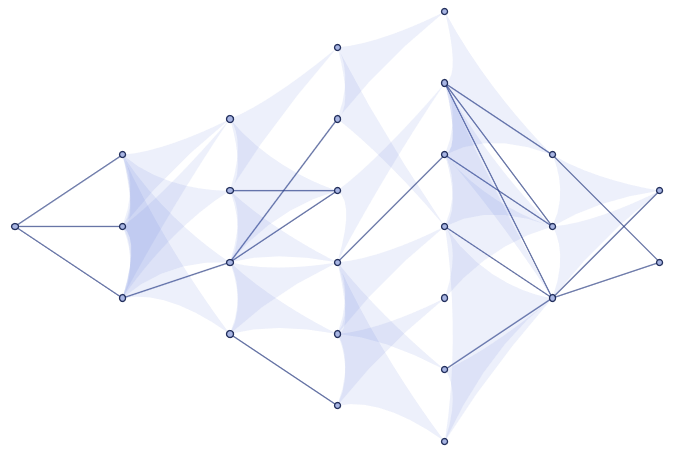}
\end{center}
This accounts for the basic infrastructure of NHNs however we need to specify how the node values will be operated at each site. Let us show this in detail for a sample case of two layers, one indexed $1$-$5$ and the subsequent $a$-$b$. Here we present an ordinary (pairwise) network on the right and a hypernetwork on the left:
\begin{center}
    \includegraphics[scale=0.6]{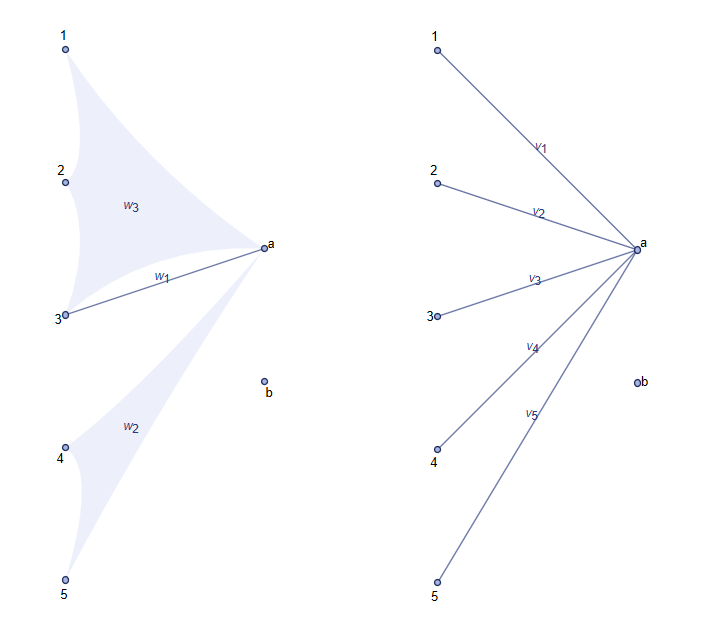}
\end{center}
where $w_i$ and $v_i$ denote the weights of each edge. If we denote by $x_1, x_2, x_3, x_4, x_5$ the scalar values at the nodes of the first layer and $x_a, x_b$ the scalar values of the subsequent layer, the hyperedge distribution (left) will imply the following operation to compute the value at the $a$ node:
\begin{equation*}
    x_a = w_1 x_3 + w_2 x_4 x_5 + w_3 x_1 x_2 x_3.
\end{equation*}
Providing a suitable activation function $\alpha_a:\mathbb{R}^+\to \mathbb{R}^+$ the value at node $a$ will be given by $\alpha_a(x_a)$. Note, in contrast, that the ordinary case of a feedforward network (right) would correspond to the following computation to be performed to determine the scalar value at the $a$ node:
\begin{equation*}
    x_a = v_1x_1 + v_2x_2 + v_3x_3 + v_4x_4 + v_5x_5,
\end{equation*}
indeed recovering the standard binary dot product between component vectors of node values and weights.

We thus see that NHNs generalize standard feedforward neural networks and present great potential for improvement on conventional models. We outline the main reasons here:
\begin{itemize}
    \item Higher-arity weights can capture multiple pair-wise weights in a single datum.
    \item Higher-arity dot products are implicitly implemented in the model at the level of the architecture.
    \item Computing node values with higher-arity weights introduces polynomial dependencies, extending the usual linear framework to a fully multilinear one.
    \item Several discrete parameters of the hypernetwork can be tuned independently and with fine control for particular applications. For instance: maximum or minimum weight arity, average incidence, arity density, etc.
\end{itemize}

\subsection{From Artificial Neural Anatomy to Artificial Neural Networks} \label{NeurAnatomy}

Having introduced the novel notion of NHN in the previous section, we shall comment here how such objects can be generated in a natural way from parsimonious higher order structures.

Our motivation comes from the observation that artificial neural networks (ANNs) are best understood as discrete representations of neural activity in living organisms. However, the underlying anatomical structures are missing from the explicit implementation of conventional ANNs. Naturally, we do not see fully sequential structures in animal brains, as the nervous tissue forms complex distributed structures that (generally) have no obvious input/output interfaces. Nevertheless, neural activity itself, i.e. the propagation of nervous impulses of electrochemical potential across neurons, has a much more direct interpretation in terms of signal throughput and thus ANNs can be interpreted as being a discrete computational model of this phenomenon.

The goal is then to define an artificial analogue of the nervous tissue where the neural activity is supported. Following our proposal from the previous section, we assume that artificial neural activity is well captured by NHNs, which were shown to be natural generalizations of feedforward artificial neural networks.

Our proposal for artificial neural anatomy is an undirected hypergraph without any sequential structure imposed on nodes, i.e. just a collection of hyperedges. We call this the \textbf{anatomical hypergraph}. The hyperedges may be boolean, effectively carrying only weights valued in the field of two elements $\{0,1\}$, or may carry arbitrary scalar values $w\in \mathbb{R}$ that represent the throughput of signals passing through them. We shall only consider the boolean case for simplicity as we illustrate the basic algorithm that takes a hypergraph and outputs a NHN. Let us consider the following small hypergraph as an example of a sparse artificial brain:
\begin{center}
    \includegraphics[scale=0.5]{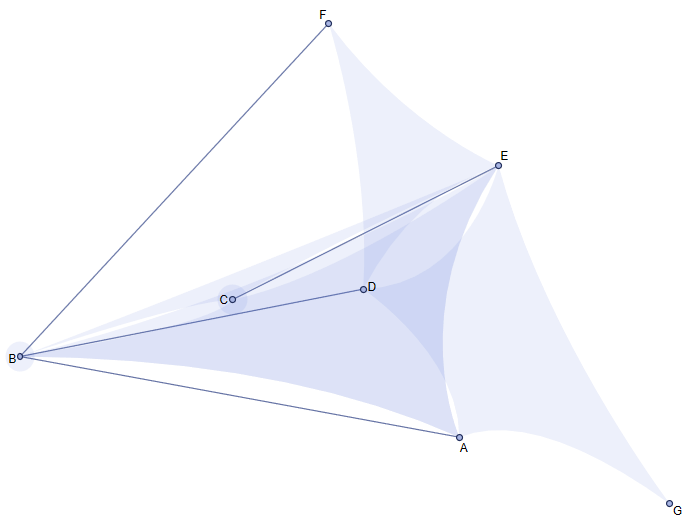}
\end{center}
Nodes are labeled with letters for ease of reference.

The basic combinatorial structure of the anatomical hypergraph is thought to be analogous to the physical distribution of neurons in natural nervous tissue with nodes representing (simplified) synaptic connections. Signals would thus propagate from nodes through hyperedges to reach other nodes. The first step in the algorithm is simply to identify a collection of nodes where the signal enters the system. These will be denoted \textbf{input nodes}. In our example let us consider nodes B and F as the input nodes. Then the signal propagates following the adjacency structure of the hypergraph reaching nodes that are connected to the input nodes and that have not been reached by the signal before. In our example, we can see that B is connected to A, C, D and E and that F is connected to E and D; therefore the first layer of nodes reached by the signal is A, C, D and E, which will be labeled with $1$ on the hypergraph. Note that B and F are themselves connected but they already carried signal in the input step. In the next step, we take layer $1$ and repeat the process, we see that the only node that is connected to some of A, C, D, E and has not been reached by the signal is G. Thus, G will be layer $2$. The final layer is also called the \textbf{output nodes}. Here is a depiction of the algorithm for our example:
\begin{center}
    \includegraphics[scale=0.5]{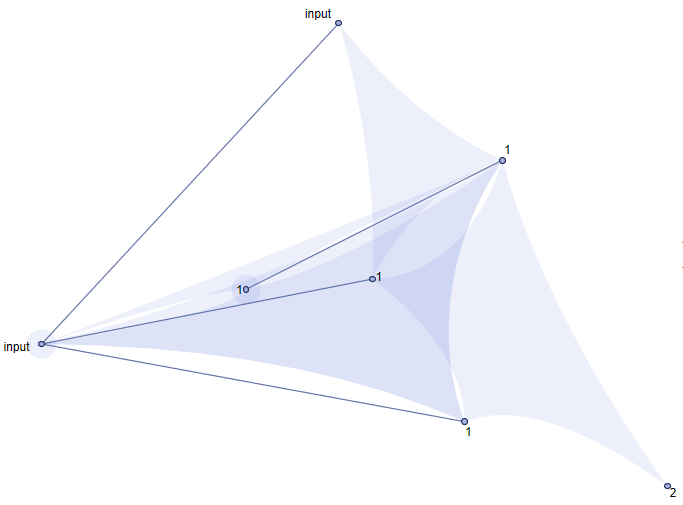}
\end{center}

This process has induced a sequential ordering of nodes into layers, thus preparing the ground for the generation of the desired NHN. To complete the process we simply take the adjacency structure of the anatomical hypergraph and define hyperedges between layers based on the following rule: take a node $y$ in layer $i$ and an incident hyperedge on it with weight $w$, then take all the nodes $\{x_1,\dots,x_n\}$ incident on that edge and that belong to layer $i-1$; then the NHN has a hyperedge of the form $\{x_1,\dots,x_n,y\}$ connecting layer $i-1$ and layer $i$. If we apply this rule to our example, we get the following boolean NHN:
\begin{center}
    \includegraphics[scale=0.6]{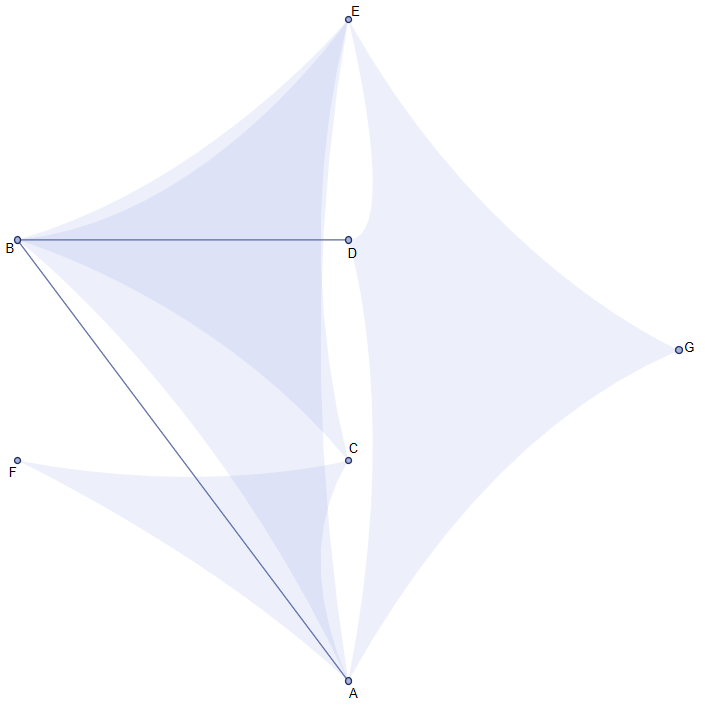}
\end{center}
Supplementing weights for each hyperedge and appropriate activation functions, we get the full structure of a neural hypernetwork.

\subsection{Summary}
In this section we have shown how using a signal propagation interpretation and biological inspiration from the observation of neural anatomy vs neural activity, allows for an anatomical hypergraph (a parsimonious non-sequential combinatorial structure) to generate all the information necessary to define ANNs, specifically in the form of neural hypernetworks. An important consideration in the deployment of hypernetworks is the combinatorial explosion that happens at lower node counts given the increased arity of the connectivity. We propose an evolutionary approach that aims to tackle this challenge in the following section.

%% file: Sections/5-Evolution.tex
\section{The Evolutionary Perspective\label{sec:Evolution}}

The introduction of higher-order structures in machine learning architecturs poses various implementation challenges. The use of hypergraphs expands the design space considerably: hyperedges allow variable arities and richer patterns of connectivity, introducing higher-order interaction motifs that yield nonlinear dependencies not reducible to simple pairwise combinations.\footnote{For example, a ternary hyperedge connecting nodes $x_1, x_2, x_3$ might map to a multiplicative interaction $f(x_1, x_2, x_3) = w \cdot x_1 x_2 x_3$. This dependency cannot be decomposed into a sum of pairwise terms (e.g., $x_1x_2 + x_2x_3 + x_1x_3$), and therefore represents a genuinely higher-order, nonlinear motif within the architecture. Concrete illustrations of such operations appear earlier in Section~3.4 (higher dot products) and Section~5.1 (higher-arity feedforward networks).} This added combinatorial complexity calls for the use of adaptive algorithms that do not rely on brute-force searches or bulk-parallel computation. Inspired by biology, we propose to take an evolutionary perspective, given its proven record in tackling computational bottlenecks in other optimization tasks.

Additionally, we note that most progress in machine learning and artificial intelligence to date has come from a combination of human algorithmic engineering and brute-force computing on ever larger data sets. The leap from recurrent neural networks to BERT, made possible by the algorithmic innovation of transformers, is an example of the former; the progress from GPT-1 to GPT-3 an example of the latter. We've noted  the distinction between neural anatomy and neural activity: modern model development mirrors these twin considerations. Engineering focuses on model architecture (anatomy), while training focuses on the content fed into models (activity). As we consider the central focus of this paper---the nature of representation in neural networks---it is natural to ask: is there an optimal neural architecture that most efficiently represents what is learned, and consequently facilitates correspondingly efficient training and inference? And if so, how do we sift through the universe of possible architectures to find it?  

Neuro-evolution refers to the use of evolutionary algorithms to optimize neural networks, typically by modifying network topologies, weights, or hyperparameters. A classical example is the NEAT algorithm (NeuroEvolution of Augmenting Topologies) \cite{stanley2002neat}, which evolves both the structure and parameters of networks simultaneously. The appeal of this approach is that it treats model design as an adaptive search process, capable of uncovering novel architectures beyond those anticipated by human engineers. This appears as very promising approach for the challenges at hand, as we shall argue in further detail in this section.

\subsection{From Anatomy and Activity to Ontogeny}  
To the twin considerations of neural anatomy and activity already discussed, evolution adds a third: neural ontogeny. In biology, ontogeny refers to the development of an organism from zygote to adult. In evolutionary computation, the analogy is the mapping from \emph{genotype} to \emph{phenotype}. In our setting, the anatomical hypergraph introduced in Section~5 can be regarded as the genotype, while the generated neural hypernetwork (NHN) is the phenotype. The algorithmic process of translating a hypergraph into a functioning NHN is the ontogeny of the model. This perspective provides the conceptual bridge to applying evolutionary methods: evolution can operate on hypergraph genotypes, whose expressed NHNs are then evaluated for fitness.  

\subsection{Evolutionary Strategies for NHNs}  
Applying evolutionary algorithms to neural hypernetworks (NHNs) requires specifying the familiar components of any evolutionary system: a genotype, a phenotype, variation operators, and a fitness evaluation. In this setting, the anatomical hypergraph serves as the genotype, with its nodes and hyperedges encoding the potential structure of a network. Each candidate hypergraph is then expressed as a phenotype through the generative procedure outlined in Section~5, guaranteeing that every genotype corresponds to a viable NHN.  

Variation is introduced at the hypergraph level. Mutations may add or remove hyperedges, alter their arities, or rewire connections, while recombination can splice together motifs from different parent graphs. In this way, evolution explores not only parameter values but also the higher-order patterns of connectivity that define NHNs.  

Once instantiated, each NHN phenotype is evaluated for fitness against task-specific objectives---accuracy on a predictive benchmark, computational efficiency, robustness to perturbation, or some weighted combination of these. Importantly, the fitness landscape can be multi-objective, balancing raw performance with other desirable properties such as sparsity or interpretability.  

The evolutionary process itself can follow different dynamics depending on the goal. Conventional selection pressure may be sufficient for optimization, while techniques such as novelty search\footnote{See, for example, Stanley and Lehman, \emph{Why Greatness Cannot Be Planned: The Myth of the Objective} (Springer, 2015), and related work on novelty search in evolutionary computation.} can be used to encourage exploration of unusual hypergraph motifs and prevent premature convergence. Together, these components form a systematic pipeline: hypergraphs as genotypes, NHNs as phenotypes, variation operators as the engine of diversity, and fitness evaluations as the guide. Through this cycle, evolution acts as a search procedure over the vast combinatorial design space of higher-order architectures.  

\subsection{Outlook}  
Evolution provides a scalable framework for navigating the enormous combinatorial space of possible NHNs, guided by fitness landscapes aligned with real-world tasks. The outcome is not merely more efficient models, but architectures whose form embodies the composable higher-order structures introduced earlier in this paper. Looking at NHNs in the light of evolution reveals them not just as static mathematical constructions, but as adaptive entities---structures that can themselves evolve toward greater functionality. In the future, just as larger data sets and more training epochs result in better models for a given architecture, so will higher mutation rates and more evolutionary epochs result in better architectures, turning an innovation problem into a search problem.

%% file: Sections/6-Conclusion.tex
\section{Future Work}
There are many intriguing directions for future research into the study of deep neural networks from the perspective of higher structures. Broadly, these directions can be classified as theoretical or applied, with theoretical topics including: (1) establishing connections to other frameworks ---  in order of highest intrigue: \cite{NeuroAlgebraicDL, GeometricDL}, and (2) developing analytical relationships between similarity scores of different arity. On the practical side, the primary challenge is developing algorithms (evolutionary or otherwise) for the intelligent exploration of the vast sea of possible higher-arity architectures, as the brute force sampling employed in \cite{PaperB} is highly inefficient. Somewhere in between is the question: do there exist meaningful metrics on such architecture spaces? 

\section{Conclusions: Towards Artificial Neuroscience}

By distinguishing between artificial neural anatomy (model structure) and artificial neural activity (computation), we open the door to a more comprehensive understanding of machine cognition. Just as cognitive neuroscience has benefited from mapping brain structures to behaviors, we can begin to correlate model architectures with capabilities and failure modes.

Inspired by the brain spatial navigation system, specifically, the discovery of grid cells and place cells, we propose that deep learning models may also develop cognitive maps. These maps would not be spatial in the geometric sense, but topological representations of conceptual domains, formed through structured tensor interactions.

To explore this idea, we propose integrating evolutionary processes and genetic programming into model design. Rather than hand-crafting architectures, we can evolve them under constraints derived from higher-order mathematical properties. This could simulate a kind of artificial neurogenesis, where structure emerges adaptively to support particular tasks or environments.

Such a system would blur the line between symbolic and neural representations, leading to models that are not only powerful, but interpretable. These models may one day explain their reasoning, adapt autonomously, and even communicate concepts in human-comprehensible terms.


In short, this project lays the groundwork for an artificial neuroscience: a unified theory of learning, representation, and reasoning based on higher-order structures in deep learning.